\documentclass[10pt,twocolumn,letterpaper]{article}

\usepackage{wacv}
\usepackage{times}
\usepackage{epsfig}
\usepackage{graphicx}
\usepackage{amsmath}
\usepackage{amssymb}

\makeatletter
\@namedef{ver@everyshi.sty}{}
\makeatother

\usepackage{pgfplots} %
\usepackage{pgfplotstable} %
\pgfplotsset{compat=newest} %
\usetikzlibrary{plotmarks} %

\usepackage[]{algorithm2e}
\usepackage{url}

\usepackage{float}
\usepackage{tabularx}
\usepackage{multirow}
\usepackage{booktabs}
\usepackage{colortbl}
\usepackage{xcolor}

\DeclareMathOperator*{\argmax}{argmax}

\wacvfinalcopy %

\def\wacvPaperID{666} %

\ifwacvfinal\fi
\begin{document}

\RestyleAlgo{boxruled}

\title{UnOVOST: Unsupervised Offline Video Object Segmentation and Tracking}

\newcommand{\footremember}[2]{%
   \thanks{#2}
    \newcounter{#1}
    \setcounter{#1}{\value{footnote}}%
}
\newcommand{\footrecall}[1]{%
    \footnotemark[\value{#1}]%
}

\author{Jonathon Luiten\footremember{cont}{Equal Contribution \newline Accepted for publication at WACV 2020} \hspace{20pt} Idil Esen Zulfikar\footrecall{cont} \hspace{20pt} Bastian Leibe\\
	Computer Vision Group, RWTH Aachen University\\
	{\tt\small \{luiten,leibe\}@vision.rwth-aachen.de \hspace{5pt} idil.esen.zuelfikar@rwth-aachen.de}
}
\maketitle
\ifwacvfinal\thispagestyle{empty}\fi

\newcommand{\PAR}[1]{\vskip0pt \noindent {\bf #1~}}
\newcommand{\PARbegin}[1]{\noindent {\bf #1~}}
\newcommand{\TODO}[1]{\textcolor{red}{TODO: #1}}

\setlength{\textfloatsep}{12pt}

\begin{abstract}
We address Unsupervised Video Object Segmentation (UVOS), the task of automatically generating accurate pixel masks for salient objects in a video sequence and of tracking these objects consistently through time, without any input about which objects should be tracked. Towards solving this task, we present UnOVOST (Unsupervised Offline Video Object Segmentation and Tracking) as a simple and generic algorithm which is able to track and segment a large variety of objects. This algorithm builds up tracks in a number stages,  first grouping segments into short tracklets that are spatio-temporally consistent, before merging these tracklets into long-term consistent object tracks based on their visual similarity. In order to achieve this we introduce a novel tracklet-based Forest Path Cutting data association algorithm which builds up a decision forest of track hypotheses before cutting this forest into paths that form long-term consistent object tracks. When evaluating our approach on the DAVIS 2017 Unsupervised dataset we obtain state-of-the-art performance with a mean $\mathcal{J}\&\mathcal{F}$ score of $67.9$\% on the \texttt{val}, $58$\% on the \texttt{test-dev} and $56.4$\% on the \texttt{test-challenge} benchmarks, obtaining first place in the DAVIS 2019 Unsupervised Video Object Segmentation Challenge. UnOVOST even performs competitively with many semi-supervised video object segmentation algorithms even though it is not given any input as to which objects should be tracked and segmented.
\end{abstract}

\begin{figure}[t!]
	\centering
	\includegraphics[width=0.8\linewidth]{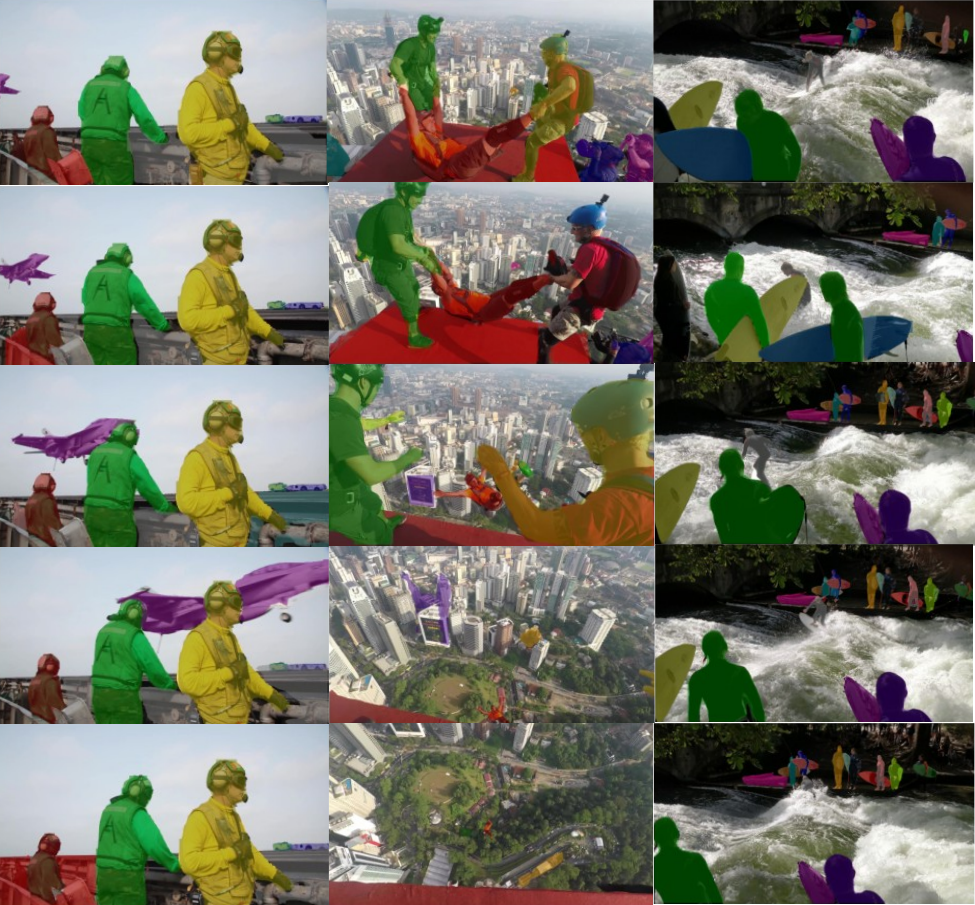}
	\caption{Example results of UnOVOST on three sequences from the DAVIS Unsupervised Dataset. UnOVOST is able to accurately segment and track many diverse objects simultaneously. Frames $1$,$\frac{T}{4}$,$\frac{T}{2}$, $\frac{3T}{4}$ and $T$ are shown, where $T$ is the number of frames in the sequence.}
	\label{fig:teaser}
\end{figure}

\vspace{-1pt}
\section{Introduction}

Video Object Segmentation (VOS) aims at automatically generating accurate pixel masks for objects in each frame of a video, then associating those proposed object pixel masks in the successive frames to obtain temporally consistent tracks. VOS has mostly been tackled in a semi-supervised fashion \cite{luiten2018premvos,voigtlaender17BMVC,Voigtlaender19CVPR}, where the object masks of the objects to be tracked in the first-frame are given, and only those objects need to be tracked and segmented throughout the rest of the video.

In this paper we tackle VOS in the more general unsupervised setting \cite{Caelles_arXiv_2019}. In such a setting we need to detect all of the possible objects in the video and track and segment them throughout the whole video. Results of our method on this task can be seen in Figure \ref{fig:teaser}. In this setting, methods are evaluating against a possibly incomplete set of ground-truth objects. As such methods are not penalized for segmenting more objects than present in the ground-truth. However, the number of predictions that can be made is limited in that predicted masks may not have overlapping pixels, and a maximum number of objects may be proposed across a whole video. As such UVOS methods must seek to segment and track the most salient objects in a video regardless of the category of those objects. Saliency here is defined as the objects that catch and maintain the gaze of a viewer across the whole of the video sequence. The definition of an object is also important and possibly ambiguous. Two important factors in determining objectness are that objects should consistently have common fate, moving together consistently throughout the scene, and that they should also be semantically consistent.

An algorithm that tackles the UVOS task has many interesting real-world applications. One such example is in robotics and autonomous vehicles where it is of crucial importance to be able to understand the precise location and motion of a huge variety of objects, from far more categories than present in any labeled dataset.

\begin{figure}[t!]
	\centering
	\includegraphics[width=\linewidth]{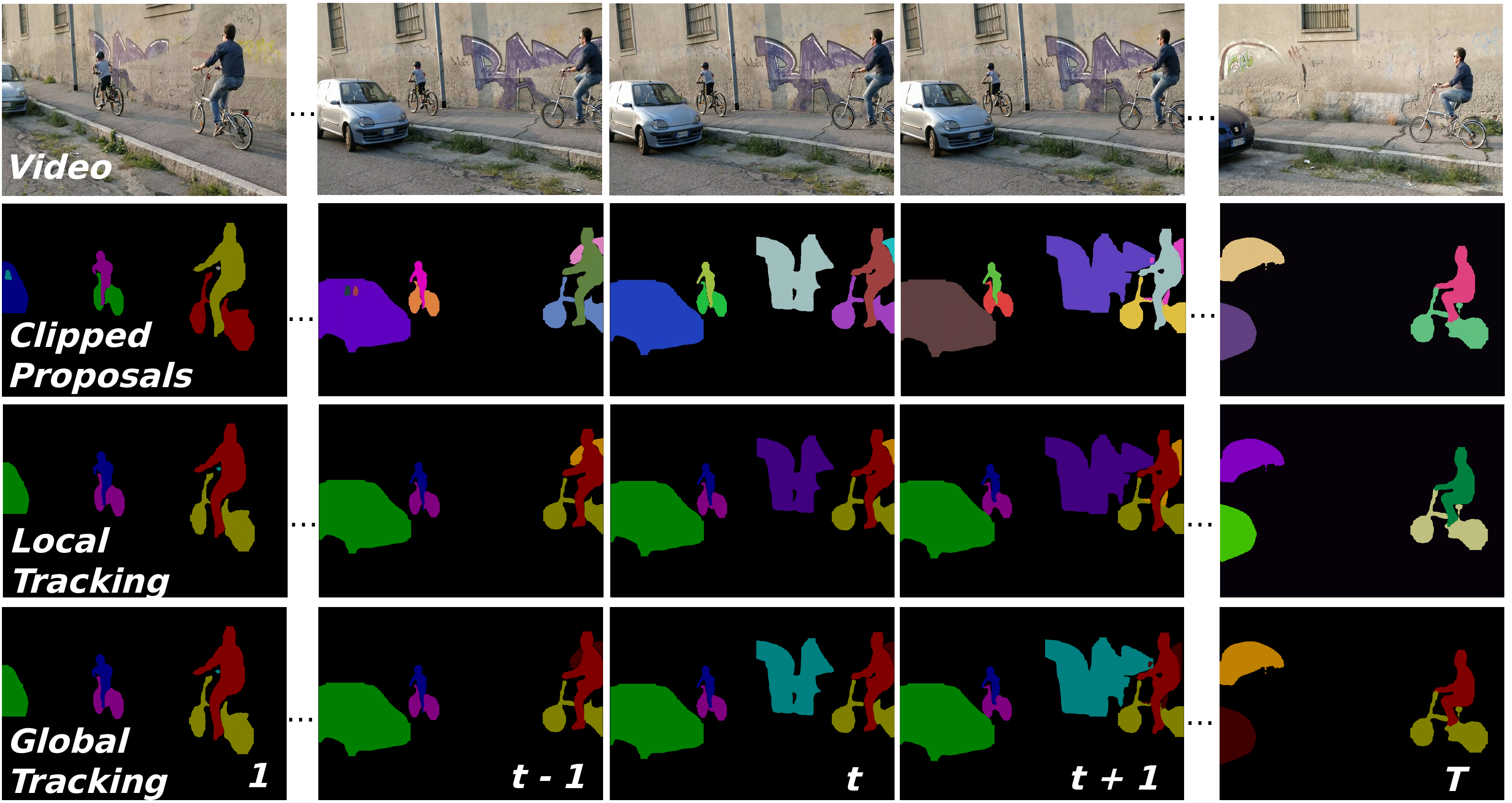}
	\caption{An overview of the UnOVOST algorithm. From an input video (row 1) a number of object mask proposals per frame are generated, sub-selected and clipped to have non-overlapping pixels (row 2). These are grouped into short-tracklets using the spatio-temporal consistency of these segments under optical-flow (row 3). These tracklets are then merged into long-term consistent object tracks using the tracklets' visual similarity and our novel Forest Path Cutting (FPC) data association algorithm.}
	\label{fig:overview}
\end{figure}

To solve this UVOS task, we present the UnOVOST (Unsupervised Offline Video Object Segmentation and Tracking) algorithm. This algorithm hierarchically builds up object segmentation tracks in multiple stages (see Figure \ref{fig:overview}). After obtaining a set of candidate object proposal masks per frame using Mask R-CNN \cite{He17ICCV}. 
We then reduce the set of mask proposals to a set which does not contain overlapping pixels by sub-selecting and clipping the given proposals. 
In order to perform segment tracking we use two main similarity cues, the spatio-temporal consistency of the mask segments in contiguous frames under optical flow warping, and the appearance-based visual similarity of the mask segments encoded as an object re-identification vector.

We then develop a novel data-association algorithm using these two similarity cues which accurately merges these mask segments into tracks. Our algorithm works in two stages. The first stage uses only the spatio-temporal consistency cues to merge segments in contiguous frames into short-tracklets which contain segments that are very likely to belong to the same object. In a second stage, we then merge these short-tracklets into long-term consistent tracks using their visual similarity. This two stage process has the benefit that the easy tracking decisions are made early and then fixed, reducing the size of the required search-space for performing data-association, and enabling information to be pooled over segments within a tracklet to better model object properties used for tracking. %

For the second-stage, we propose a novel Forest Path Cutting (FPC) algorithm. This algorithm builds a forest consisting of decision trees of possible track hypotheses. The final set of object tracks is then produced by iteratively cutting paths from this forest, until the forest is divided into a non-conflicting set of paths which are the final tracks. This algorithm is both simple and efficient while being powerful enough to model the combinatorial complexity of the long-term data-association problem.

When evaluating UnOVOST on the unsupervised DAVIS benchmark dataset \cite{Caelles_arXiv_2019} we achieve state-of-the-art results compared to all previous methods, as well as results competitive with semi-supervised methods using the given first-frame mask as guidance for which objects to track and segment. Our method also achieves the first place in the DAVIS 2019 Unsupervised Video Object Segmentation Challenge. When extending our method to the task of Video Instance Segmentation (VIS) by adding classifying our object tracks, we also obtain state-of-the-art results on the YouTube-VIS benchmark and won the 2019 YouTube-VIS challenge.

\section{Related Work}

\PAR{Multi-Object Unsupervised Video Object Segmentation.}
The UVOS task (also known as zero-shot VOS) is quite recent, and there are few methods that tackle this task. UVOS \cite{Caelles_arXiv_2019} was proposed as a challenge task for the 2019 DAVIS Challenge on Video Object Segmentation. \cite{Caelles_arXiv_2019} evaluate the RVOS (Recurrent Video Object Segmentation) \cite{rvos} method for the UVOS task. This method uses a number of recurrent neural networks, one over the set of objects, and one over time to generate  tracks. Our method, UnOVOST, outperforms RVOS by more than 25 percentage points on the $\mathcal{J}\&\mathcal{F}$ metric on all benchmarks.
In the 2019 DAVIS Challenge, our method obtained first place. The second \cite{2nd} and third \cite{3rd} place methods presented very different approaches to the UVOS task. \cite{2nd} propose to run a detector on each frame, as well as a series of single object trackers used to merge the detections into tracks. \cite{3rd} adapts \cite{luiten2018premvos} from semi-supervised VOS to UVOS task, while adding a proposal pruning step after a number of initial frames, and then tracking these objects as though this was a semi-supervised task.

\PAR{Single-Object Unsupervised Video Object Segmentation.}
There has been a number of papers tackling single-object unsupervised video object segmentation (SOUVOS) \cite{Li_2018,Li_2018_ECCV,Jain_2017,Tokmakov_2017,Tokmakov_2_2017}. This is inherently a different problem to the multi-object task that we tackle in this paper. SOUVOS is closer related to foreground/background segmentation as it requires only one foreground area to be segmented which often is a grouping of multiple objects into one foreground object. This task is often evaluated on the DAVIS 2016 single object benchmark \cite{davis2016}. This task requires estimating the single most salient grouping of foreground objects in a video. Methods that tackle this task, such as \cite{Jain_2017} and \cite{Tokmakov_2_2017} often perform two class segmentation on an image concatenated with optical-flow.

\PAR{Motion Segmentation.}
Another related field is motion segmentation. This task differs from video object segmentation in that it only requires the segmentation of objects that are moving \cite{rubrik}, whereas UVOS requires the segmentation of all objects whether they are moving or not. Motion segmentation methods \cite{bideauCVPR18,xie2018object,dave2019segmenting} are often based on low-level vision features such as the optical-flow. \cite{dave2019segmenting} adapts Mask R-CNN \cite{He17ICCV} to operate on both image and optical-flow input. \cite{xie2018object} extract features from the combination of the image and the optical-flow and clusters these. \cite{bideauCVPR18} develops a two-stage model that estimates piece-wise rigid motions, which are then merged into objects. This is evaluated as either a multi-object task, or a foreground/background estimation task often using the FSMB \cite{OB14b} dataset.

\PAR{Multi-Object Semi-Supervised Video Object Segmentation.}
Semi-Supervised Video Object Segmentation (SSVOS) is where the objects that need to be tracked are given as segmentation masks in the first frame. Algorithms that tackle this task often finetune a segmentation network on the given first frame \cite{OSVOS,voigtlaender17BMVC,luiten2018premvos}, or propagate from the given first-frame mask directly to the rest of the video \cite{Voigtlaender19CVPR,Oh18CVPR}. These methods are not able to be easily adapted to UVOS as they rely heavily on the first-frame mask. \cite{luiten2018premvos} is the closest related SSVOS to our method, as it also produces generic object segmentation proposals and links these in time with spatio-temporal and visual similarity cues. However, unlike our method, \cite{luiten2018premvos} finetunes all of its components heavily on the first-frame, uses the given-first frame to guide which objects to track, and performs data association in a simple frame-by-frame fashion. SSVOS is often evaluated on the DAVIS 2017 semi-supervised dataset.

\PAR{Video Instance Segmentation and Multi-Object Tracking and Segmentation.}
Recently, the related tasks of Video Instance Segmentation (VIS) \cite{VIS} and Multi-Object Tracking and Segmentation (MOTS) \cite{mots} has been proposed. These tasks are similar to UVOS in that objects need to be tracked and segmented without being given guidance on which particular instances are to be tracked. However, these tasks differs from UVOS in that only objects belonging to a specified categories need to be tracked and segmented, as well as these being classified correctly. This significantly simplifies the task, and limits the applicability of methods that tackle these tasks. MOTS differs from VIS in that in MOTS sequences are much longer and many more instances are present with objects disappearing and reappearing much more often. MOTS is evaluated on the KITTI and MOTChallenge datasets \cite{mots}. VIS on the YouTubeVIS benchmark \cite{VIS}. We extend UnOVOST from the UVOS task to the VIS task by classifying our resulting tracks, and also achieve state-of-the-art performance on this task.

\PAR{Category Agnostic Multi-Object Tracking.}
Previously, a number of methods \cite{camot,Osep19ICRA} have attempted to extent multi-object tracking methods beyond tracking objects from a predefined set of object categories. These methods \cite{camot,Osep19ICRA} have typically relied on the presence of stereo-camera input to obtain 3D information for evaluating the objectness of generic object proposals. These methods also make no attempt to create a set of object tracks without overlapping segment masks, instead they create a large set of track proposals that have large overlap with one-another, often tracking the same object multiple times on different scales. Our method by contrast works on monocular video and creates a set of segmentation tracks without overlapping segment masks.

\begin{figure}[t!]
	\centering
	\includegraphics[width=0.8\linewidth]{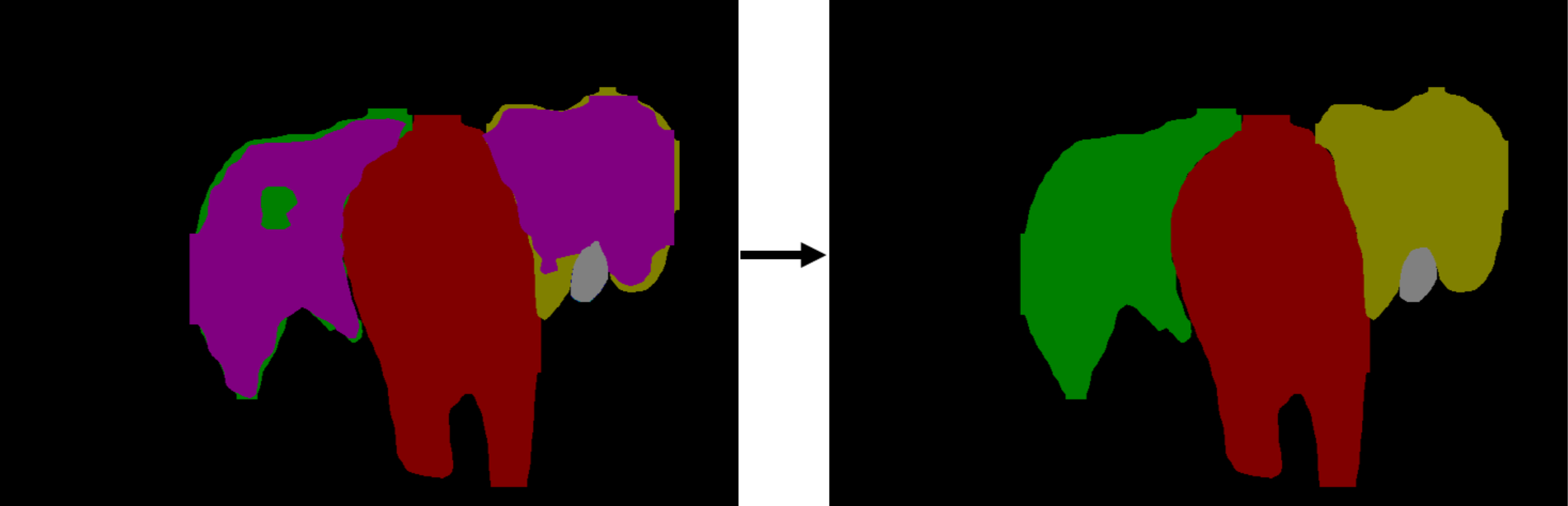}
	\caption{Initial stage of the UnOVOST algorithm. Mask proposals are generated by Mask R-CNN with a low scoring threshold that generates a large number of overlapping proposals. These are then sub-selected and clipped based on their score and intersection to produce a set of non-overlapping mask proposals in each frame.}
	\label{fig:overlap}
\end{figure}

\PAR{Data-Association for Multi-Object Tracking.}
The task of multi-object tracking (MOT) has a long research history \cite{leal2017tracking}. The leading paradigm for MOT has become tracking-by-detection, where a set of object detections are proposed, and tracking is reduced to a data-association problem. In this paper we propose a new data-association algorithm designed specifically for the UVOS task. Previous data-association methods are either too simple or unnecessarily complex. Many methods such as those used in \cite{mots} and \cite{luiten2018premvos} only take into account associations in the previous frame, or the previous and first frame \cite{luiten2018premvos}, and do not use the context from the whole video. On the other hand, data-association algorithms such as \cite{mht} are unnecessarily complex in that they produce an exponentially large number of potential track hypotheses and score each of these individually. Our data-association algorithm is able to take advantage of a number of simplifications to be able to use the whole video context to evaluate the likelihood of a tracking hypothesis, while being much simpler and efficient. We take advantage of the fact that mask segments can not overlap to significantly reduce the set of possible tracking combinations. Furthermore, we split tracking into two components, firstly grouping proposals based on spatio-temporal consistency, before only using visual similarity in a second stage. These simplifications in combination with our efficient Forest Path Cutting algorithm, results in an algorithm that is accurate, powerful and efficient.

\section{Approach}
In this section we detail the specifics of our novel UnOVOST algorithm for tackling the UVOS task. Our method begins by generating a large set of generic object proposals and sub-selects and clips these to be non-overlapping. These proposals are then grouped over contiguous frames into short-tracklets based on the spatio-temporal consistency of the object proposals under an optical-flow warping. We then merge these tracklets into long-term consistent object tracks using our novel Forest Path Cutting algorithm and visual similarity cues between tracklets. Finally, a final set of object tracks is selected based on their video saliency.

Unlike previous approaches \cite{luiten2018premvos,mots,dave2019segmenting} our algorithm is able to segment and track objects regardless of their object class, whether they are static or undergo motion, and whether they are present in the foreground or background of the scene. Our algorithm instead relies on a more general concept of objectness to determine what should consist of an object to be tracked and segmented. An overview can be seen in Figure \ref{fig:overview}.

\begin{algorithm}[!t]
\footnotesize
\caption{Forest Path Cutting Algorithm (FPC)}
\label{alg:FPC}

\KwData{Tracklets $L_i$ with average ReID vectors $R_i$, beginning \\ \hspace{-0.5pt} timestep $b_i$ and ending timestep $e_i$, ordered by increasing $b_i$. }
\KwResult{Tracks $F_j$ which are groupings of tracklets.}

\textbf{Define:}Visual Similarity $V_{i,j}$
$ V_{i,j} := 1 - \frac{\Vert R_i - R_j \Vert}{\max_{m \in \{1 \dots T \} ,n \in \{ 1 \dots T \} }{\Vert R_m - R_n \Vert}} $ \\

\textbf{Part 1:} Build a forest of track hypotheses, by calculating optimal predecessors $M_i$ for each tracklet $L_i$.\\
\For{$i \in \{1 \dots \vert L \vert \} $}{
\eIf{$\{j \, | \, e_{j} < b_i\} \neq \emptyset $}
{$k := \argmax\limits_{j | e_{j} < b_i}\{ V_{i,j} \}$ \\
\While{$ \{j \, | \, e_j < b_i  , \, b_j > e_k  , \, M_j = L_k \} \neq \emptyset $}{
$l := \argmax\limits_{j | e_{j} < b_i , j \neq k}\{ V_{i,j} \} $ \\
\eIf{$ l \in \{j \, | \, e_j < b_i  , \, b_j > e_k  , \, M_j = L_k \} $}
{$k := l$}
{break}
}
$ M_i := L_k $
}
{ $M_i := \emptyset $}
}

\textbf{Part 2:} Define the set of track hypotheses $H$ as the paths from root nodes to leaf nodes through the hypothesis forest, and calculate a score $C_i$ for each path. Select final tracks $F$ by iteratively cutting the optimal paths from the forest.\\
$ H := \{ \{ L_i, M_i, M_{j|L_j=M_i},\dots, L_{k | M_k = \emptyset} \} | L_i \neq M_m \forall m \}$ 

$ F:=\emptyset$\\
\While{$H \neq \emptyset $}{
\For{$H_i \in H$}{
$ C_i^V := \min\limits_{m,n | L_m \in H_i, L_n \in H_i}\{V_{m,n}\}$ \\
$ C_i^T := \sum\limits_{j | L_j \in H_i}{e_j-b_j+1} $ \\
$ C_i := 0.1 C_i^V + 0.9 C_i^T $ \\
}
$ k:=\argmax\limits_{i | H_i \in H}{C_i}$ \\
$ F := F \cup \{H_k\} $\\
\For{$H_i \in H \setminus \{H_k\}$}{
$ H_i:=H_i \setminus H_k $ \\
}
$ H := H \setminus \{H_k\} $
}

\end{algorithm}

\PAR{Object Mask Proposal Generation.}
We generate a large number of proposals, segmentation masks which cover potential objects, for a diverse range of objects. Specifically we use a Mask R-CNN \cite{He17ICCV} implementation by \cite{wu2016tensorpack} with a ResNet101 \cite{He_2016} backbone trained on COCO \cite{lin2014microsoft}. Although this network has been trained to detect the 80 COCO categories, we find that when using a low-confidence threshold this network produces adequate mask proposals for objects beyond these 80 categories. This network produces masks, bounding boxes, object categories and confidence scores for object proposals as outputs. We discard the object categories and treat all detections as if they come from the same foreground object category. We extract all proposal masks with a confidence score greater than $0.1$. An example of generated overlapping proposals can be found in Figure \ref{fig:overlap}.

\begin{figure*}[t!]
	\centering
	\includegraphics[width=0.8\textwidth]{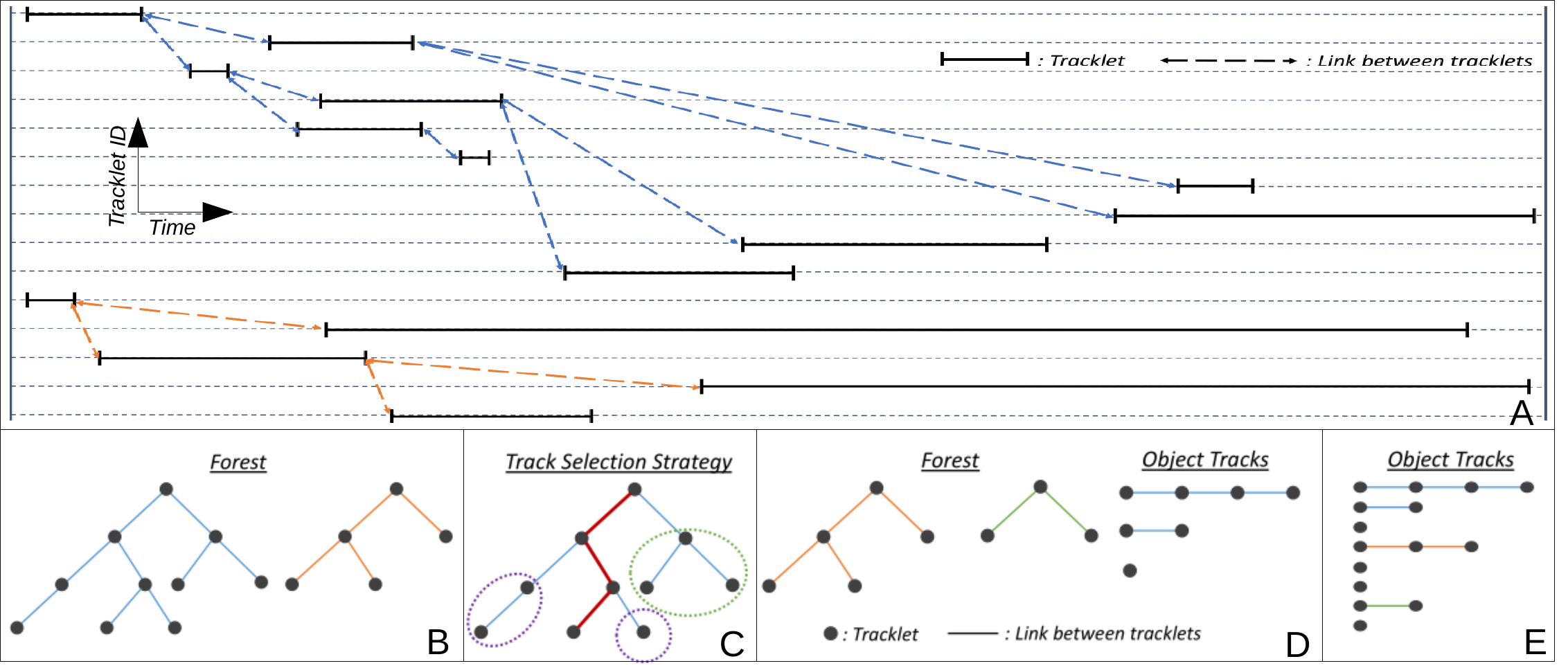}
	\caption{Visual representation of our Forest Path Cutting (FPC) algorithm. In box A, tracklets are visualized as black lines showing their temporal extent on the horizontal axis. The optimal predecessors for each tracklet are shown as blue and orange dotted lines. Box B shows that the set of tracklets with predecessors from box A defines a forest of track hypotheses. Box C shows an optimal path (in red), selected from the forest and added to the final object tracks.This optimal path is cut from its tree, dividing it into a number of sub-trees (green and purple circles). Box D shows the resulting new forest produced with sub-trees that only contain a single path added to the list of final tracks. Box E shows the final result after this process is iterated until the forest is completely divided into a set of tracks.}
	\label{fig:tracklets}
\end{figure*}

\PAR{Proposal Sub-Selection and Clipping.}
In order to simplify the tracking and segmentation problem, we initially ensure that our set of segment masks do not overlap in each frame individually before tracking these segments throughout the video. All proposal masks in a frame are compared against one another using their intersection over union (IoU) to detect overlaps. If the IoU between two proposal masks is higher than $0.2$, then the proposal mask with higher confidence score is kept and the other proposal mask is removed. This is a form of mask-based non-maximum suppression. For all the remaining masks, we clip overlaps so that the mask with the highest confidence score is on top of a mask with a lower score. This set of proposals without overlaps has three advantages, it reduces the number of proposals that need to be tracked, simplifies the matching using temporal consistency cues as conflicting proposals are removed, and removes many spurious masks not belonging to real objects. See Figure \ref{fig:overlap}.

\PAR{Tracklet Generation.}
A tracklet is a series of proposals in contiguous frames  which have been merged to belong to the same object identity.
In the first stage of our tracking algorithm we join proposals in contiguous frames into tracklets if they have a very high spatio-temporal consistency.

The spatio-temporal consistency score between two proposals in contiguous timesteps is calculated as the IoU between the proposal projection from the earlier frame and the proposal in the later frame. The proposal projection is the segmentation mask generated by warping a proposal by its corresponding optical flow vectors calculated using PWCNet \cite{sun2017pwc}. Effectively the projection of this proposal into the next timestep.

This first stage proceeds frame by frame. For each pair of contiguous timesteps, we create a complete bipartite graph whose nodes are the proposals in successive frames and whose edge scores are the spatio-temporal consistency scores. Edges are dropped from the graph if their score is less than $0.05$. We then solve the matching problem between the two sets of nodes using the Hungarian matching algorithm, which finds an optimal set of matches between the two frames. If any proposal is not matched this ends a tracklet. Tracklets may span only a single frame. 

\PAR{Merging Tracklets into Tracks.}
A track is a set of proposals over an entire video which belong to the same object identity. A track often contains multiple tracklets with potentially frames in-between them without proposals.

The second stage of UnOVOST merges tracklets into long term tracks. To do this we introduce a novel Forest Path Cutting (FPC) algorithm. An overview of this algorithm can be seen in Algorithm \ref{alg:FPC} and Figure \ref{fig:tracklets}. This algorithm merges tracklets based on visual similarity cues.
Note that spatio-temporal consistency now provides very little further value for data-association. If the tracklets could be easily determined to belong together by spatio-temporal consistency they would have been merged in the first stage. The remaining data association decisions are more difficult such as tracking objects through heavy or total occlusion. 

For each proposal we calculate a ReID vector, a representation of the visual appearance of a proposal which can be used to compare the visual similarity of proposals or tracklets, and thus to re-identify a proposal or tracklet as belonging to a certain object identity.
To calculate these vectors we use an appearance embedding network \cite{Osep19ICRA} which extracts an embedding from an image crop. 
This network is inspired from the person re-identification community. This is a wide ResNet \cite{Wu_2019} trained with a batch-hard soft-margin version of the triplet loss. This is pretrained to distinguish classes on COCO \cite{lin2014microsoft}, before being trained to distinguish instances on YouTube-VOS \cite{xu2018youtube}. It is trained so that the embedding for instances in the same track are pulled closer together in embedding space that for difference tracks.
We average ReID vectors over all proposals in a tracklet to achieve a more robust appearance representation.

To compare two tracklets we define a visual similarity score as the L2 distance of the two ReID vectors normalised to between 0 and 1, with 1 being identical, and 0 being the maximum distance between all tracklets in a video. We subtract this from one to convert it to a similarity score.

To perform long-term tracking we enumerate a set of potential track hypotheses as different combinations of tracklets. A final set of tracks can be selected as a valid subset of this set of track hypotheses.

For this task we introduce our Forest Path Cutting (FPC) algorithm as can be seen in Algorithm \ref{alg:FPC} and Figure \ref{fig:tracklets}. Initially (Part 1 and Box A) our algorithm builds up a forest of potential tracking hypotheses throughout the video by determining an optimal predecessor for each tracklet. 
Our FPC algorithm draws parallels to dynamic programming, as we wish to determine an optimal back-pointer for each tracklet to a previous predecessor tracklet. However, a naive implementation of a dynamic programming algorithm would not be able to take advantage of the desirable properties of a UVOS solution. 

To calculate optimal back-pointers, our algorithm iterates over the tracklets in order from the earliest to the latest starting time. For the current tracklet $L_i$, if there are any tracklets ending before the tracklet's start time, it determines the most similar predecessor tracklet $L_k$ based on the visual similarity score. This is an initial guess for the best predecessor tracklet, however this may belong to the same object, but not be the direct predecessor if there is another tracklet between the two that also belongs to the same object. 
We check if there are any compatible tracklets between $L_i$ and $L_k$, which have tracklet $L_k$ as their predecessor. Choosing one of these tracklets as the predecessor would result in $L_k$ still being an earlier predecessor. However, we only wish to choose one of these tracklets if it is the most visually similar tracklet to tracklet $L_i$ (except for $L_k$). We repeat this procedure iteratively until there are no more tracklets between the $L_i$ its current predecessor $L_k$, or another tracklet which does not have $L_k$ as its predecessor is the most visually similar tracklet not in the current set of predecessors.

\begin{figure}[t!]
	\centering
	\includegraphics[width=0.8\linewidth]{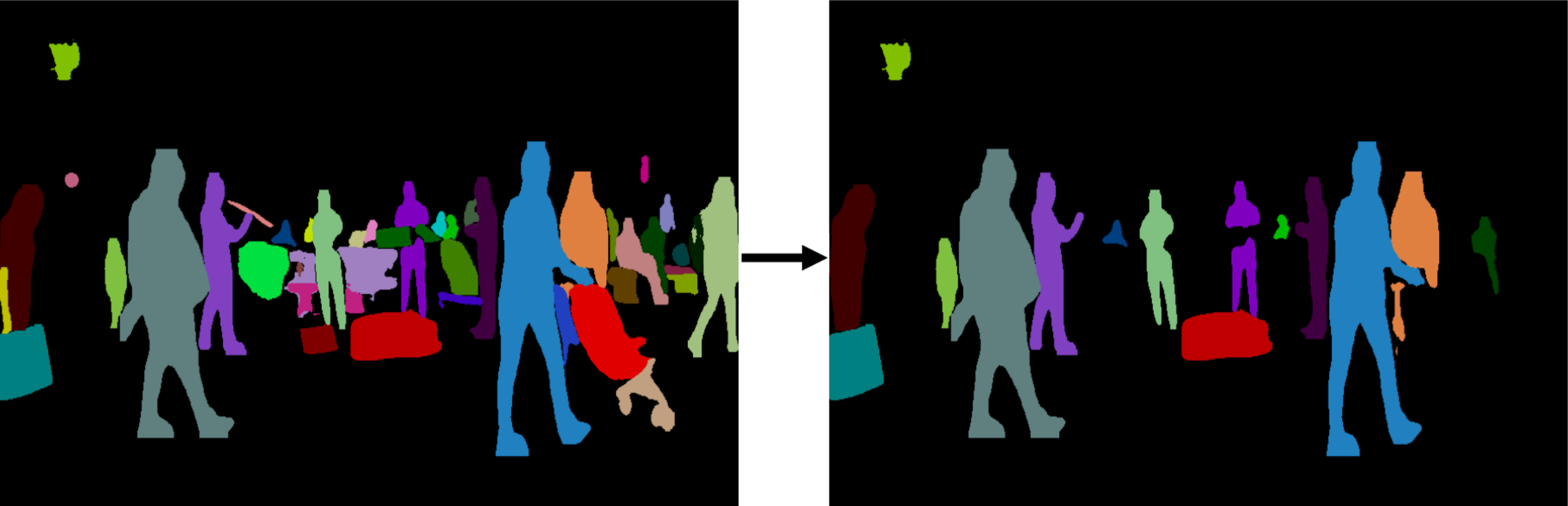}
	\caption{An example of the final stage of UnOVOST, where the final set of tracks is reduced to a maximum of 20 objects over the whole video using our video object saliency metric. Note that objects that would capture an observer's attention are retained while the rest are discarded.}
	\label{fig:salient}
\end{figure}

We now have a forest of track hypotheses, each tracklet may only have a single predecessor, but a tracklet can be the predecessor for multiple successor tracklets. The resulting forest has at least one tree whose root node corresponds to one of the tracklets with the earliest starting time. Each path $H_i$ through the trees from a root node to a leaf node in this forest is a possible long-term object track. 
We now cut this forest into a set of paths which is the best possible set of object tracks by applying a greedy recursive track selection strategy. This selects an optimal path from the forest, which is then added to the set of final long-term object tracks $F$. This path is then cut from forest, with all nodes belonging to this path being removed from the forest, and the forest rearranging itself into a new set of trees with the remaining nodes. This can be seen in Algorithm \ref{alg:FPC} part 2 and Figure \ref{fig:tracklets} parts B-E. 
To score paths we use a combination of a Visual Consistency Score $C_i^V$ and a Temporal Density Score $C_i^T$.

The visual consistency score is the minimum visual similarity embedding distance between any two tracklets in a path.
The temporal density score is the fraction of frames of a video where there is a segment present. This penalizes large temporal gaps between tracklets, making it more likely that objects undergoing short occlusion are correctly tracked, and ensures that the most salient objects are grouped consistently throughout the video, as objects to be tracked in UVOS are present in mostly all frames.
The final path score is a weighted sum of 90\% the temporal density and 10\% the appearance consistency. 

We select paths with the highest score through the forest, add these to a final list of tracks $F$, and cut these from the current forest, reshaping the forest into a new set of trees. This algorithm select a set of object tracks which do not include any overlapping tracklets and have long-term temporal consistency.

\newcommand{\mc}[1]{\shortstack{#1}}
\begin{table}[t!]
	\scriptsize
 	\newcommand{\mmr}{\arrayrulecolor{lightgray}\cmidrule[0.25pt]{2-6}\arrayrulecolor{black}}
 	
 	\setlength{\tabcolsep}{6pt}
 	\begin{center}
 		\begin{tabularx}{\textwidth}{cccccccc}
 			\cmidrule[1pt]{1-7}
 			\multicolumn{3}{c}{} & Ours & VSD \cite{2nd} & KIS \cite{3rd} & RVOS \cite{rvos}\\	
 			
 			\cmidrule[1pt]{1-7}
 			
 			\multirow{7}{*}{\mc{U17 \\ T-C}} & $\mathcal{J}\&\mathcal{F}$ & Mean & \textbf{56.4} & 56.2 & 51.6 & -\\
 			\mmr
 			& \multirow{3}{*}{$\mathcal{J}$} & Mean & 53.4  & \textbf{53.5} & 48.7 & -\\
 			& & Recall & 60.9 & \textbf{61.3}  & 55.1  & -\\
 			& & Decay & 1.5  & \textbf{-2.1} & 4.0 & - \\
 			\mmr
 			& \multirow{3}{*}{$\mathcal{F}$} & Mean &  \textbf{59.4} & 59.0 & 54.5& - \\
 			& & Recall &  \textbf{64.1} & 63.2 & 59.4 & -\\
 			& & Decay & 5.8 & \textbf{0.1} & 7.7 & -\\
 			
 			\cmidrule[0.5pt]{1-7}
 			
 			\multirow{7}{*}{\mc{U17 \\ T-D}}  &  $\mathcal{J}\&\mathcal{F}$ & Mean & \textbf{58.0} & 56.5  & 54.2 & 22.5\\
 			\mmr
 			& \multirow{3}{*}{$\mathcal{J}$} & Mean & \textbf{54.0} & 51.7 & 50.0 & 17.7\\
 			& & Recall & \textbf{62.9} & 59.9  & 58.9 & 16.2 \\
 			& & Decay & 3.5 & 21.7 & 8.4 & \textbf{1.6} \\
 			\mmr
 			& \multirow{3}{*}{$\mathcal{F}$} & Mean & \textbf{62.0} & 61.4  & 58.3 & 27.3\\
 			& & Recall & \textbf{66.6} & 65.7 & 62.1  & 24.8\\
 			& & Decay &  6.6 & 15.7 & 11.4 & \textbf{1.8} \\
 			
 			\cmidrule[0.5pt]{1-7}
 			
 			\multirow{7}{*}{\mc{U17 \\ Val}} & $\mathcal{J}\&\mathcal{F}$ & Mean & \textbf{67.9} & 56.6 & 59.9 &41.2 \\
 			\mmr
 			& \multirow{3}{*}{$\mathcal{J}$} & Mean & \textbf{66.4} & 51.7 & -  & 36.8 \\
 			& & Recall & \textbf{76.4} & - & - & 40.2\\
 			& & Decay & \textbf{-0.2} & - & - & 0.5\\
 			\mmr
 			& \multirow{3}{*}{$\mathcal{F}$} & Mean & \textbf{69.3} & 61.4 & -  & 45.7 \\
 			& & Recall & \textbf{76.9} & - & -  & 46.4\\
 			& & Decay & \textbf{0.01} & - & -  & 1.7\\ 			
 			
 			\cmidrule[1pt]{1-7}
 			
 		\end{tabularx}
 	\end{center}
 	\caption{Our results compared to all other UVOS methods on the DAVIS 2017 unsupervised benchmarks: \texttt{test-challenge} (U17 T-C), \texttt{test-dev} (U17 T-D), and \texttt{val} (U17 Val). VSD \cite{2nd} and KIS \cite{3rd} obtained second and third place (after our method) in the 2019 DAVIS Unsupervised VOS Challenge.} 
 	\label{table:results}
\end{table}

\PAR{Final Tracks Selection.}
In UVOS algorithms are not penalized for making predictions that do not overlap with ground-truth. However, the number of total object tracks that can be predicted in still limited. In the DAVIS 2017 unsupervised benchmark this is limited to $20$ predictions over the whole video. 

UnOVOST predicts a potentially large number of tracks, therefore a final object video saliency estimation step is performed to estimate the $20$ most salient objects over the whole video to report for evaluation. 

Our video saliency score $S_{sal,i}$ is calculated for each track using each tracklet $t_j$ in the track $i$: 
 \begin{align} 
  S_{sal,i} &= \sum_{j} temp(t_j) \; conf(t_j)
 \end{align}
 where $temp(t_j)$ is the temporal length of tracklet $j$ and $conf(t_j)$ is the average of the confidence scores of proposals in tracklet $t_j$. This video saliency metric prefers tracks that are present in many frames and that have high objectness confidence. An example of this is shown in Figure \ref{fig:salient}.

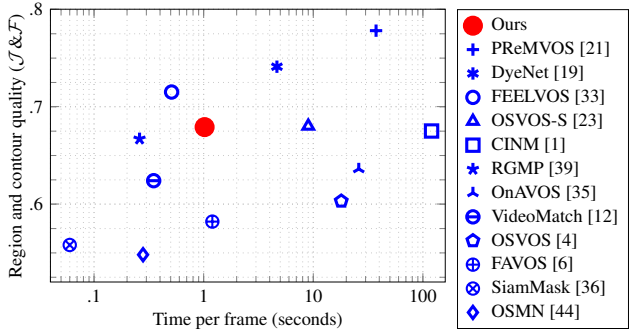
\begin{figure}
\centering
\resizebox{\linewidth}{!}{\begin{tikzpicture}[/pgfplots/width=1\linewidth, /pgfplots/height=0.75\linewidth, /pgfplots/legend pos=south east]
    \begin{axis}[ymin=0.52,ymax=0.80,xmin=0.04,xmax=160,enlargelimits=false,
        xlabel=Time per frame (seconds),
        ylabel=Region and contour quality ($\mathcal{J}$\&$\mathcal{F}$),
		font=\small,%
        grid=both,
		grid style=dotted,
        xlabel shift={-2pt},
        ylabel shift={-5pt},
        xmode=log,
        legend columns=1,
        minor ytick={0,0.025,...,1.1},
        ytick={0,0.1,...,1.1},
	    yticklabels={0,.1,.2,.3,.4,.5,.6,.7,.8,.9,1},
	    xticklabels={0.01,.1,1,10,100},
        legend pos= outer north east,
        legend cell align={left}
        ]

	    \addplot[red,mark=*,only marks,line width=0.75, mark size=4.5] coordinates{(1.02,0.679)};
        \addlegendentry{\hphantom{i}Ours}
        
		\addplot[blue,mark=+,only marks,line width=1.5, mark size=3] coordinates{(37.4,0.778)};
        \addlegendentry{\hphantom{i}PReMVOS \cite{luiten2018premvos}}
        
        \addplot[blue,mark=asterisk, only marks, line width=1.5, mark size=3]coordinates{(4.66,0.741)};
        \addlegendentry{\hphantom{i}DyeNet \cite{Li18ECCV}}
        
        \addplot[blue,mark=o,only marks,line width=1.5, mark size=3] coordinates{(0.51,0.715)};
        \addlegendentry{\hphantom{i}FEELVOS \cite{Voigtlaender19CVPR}}
        
        \addplot[blue,mark=triangle, only marks, line width=1.5, mark size=3] coordinates{(9,0.68)};
        \addlegendentry{\hphantom{i}OSVOS-S \cite{Maninis18TPAMI}}
        
        \addplot[blue,mark=square, only marks, line width=1.5, mark size=3] coordinates{(120,0.675)};
        \addlegendentry{\hphantom{i}CINM \cite{Bao18CVPR}} 
        
        \addplot[blue,mark=star, only marks, line width=1.5, mark size=3] coordinates{(0.26,0.667)};
        \addlegendentry{\hphantom{i}RGMP \cite{Oh18CVPR}}
        
         \addplot[blue,mark=Mercedes star,only marks,line width=1.5, mark size=3] coordinates{(26,	0.636)};
        \addlegendentry{\hphantom{i}OnAVOS \cite{voigtlaender17BMVC}}
        
        \addplot[blue,mark=halfcircle,only marks,line width=1.5, mark size=3] coordinates{(0.35,0.624)};
        \addlegendentry{\hphantom{i}VideoMatch \cite{Hu18ECCV}}
        
        \addplot[blue,mark=pentagon, only marks, line width=1.5, mark size=3] coordinates{(18,0.603)};
        \addlegendentry{\hphantom{i}OSVOS \cite{OSVOS}}
        
        \addplot[blue,mark=oplus, only marks, line width=1, mark size=3] coordinates{(1.2,0.582)};
        \addlegendentry{\hphantom{i}FAVOS \cite{Cheng18CVPR}}  
        
        \addplot[blue,mark=otimes, only marks, line width=1, mark size=3] coordinates{(0.06,0.558)};
        \addlegendentry{\hphantom{i}SiamMask \cite{Wang19CVPR}}       
        
        \addplot[blue,mark=diamond, only marks, line width=1.5, mark size=3] coordinates{(0.28,0.548)};
        \addlegendentry{\hphantom{i}OSMN \cite{Yang18CVPR}}

    \end{axis}
\end{tikzpicture}}
   \caption{Quality versus timing plot comparing UnOVOST to state-of-the-art semi-supervised methods on DAVIS17 \texttt{val}. All methods other than ours are ``semi-supervised" and use the given first-frame ground-truth. Our methods obtains similar results while working in an ``unsupervised" manner without using any given information about which objects should be tracked.}
   \label{fig:speedplot}
\end{figure}

\vspace{-3pt}
\section{Experiments}
\vspace{-3pt}
\PAR{Unsupervised VOS Evaluation.}
We evaluate UnOVOST on the DAVIS 2017 Unsupervised dataset \cite{Caelles_arXiv_2019}. This contains videos in four sets, with 60 \texttt{train}, 30 \texttt{val}, 30 \texttt{test-dev} and 30 \texttt{test-challenge} videos. The \texttt{train} and \texttt{val} sets contain the same videos as the DAVIS 2017 semi-supervised dataset, however they have been re-annotated according to the definition of the UVOS task. The \texttt{test-dev} and \texttt{test-challenge} sets contain new videos. All of these datasets include multiple objects per video sequence. Methods are ranked on the $\mathcal{J}\&\mathcal{F}$ metric which is the average of an area overlap ($\mathcal{J}$) and a boundary overlap ($\mathcal{F}$) metric. More details can be found in \cite{davis2016}.

\begin{figure}[t!]
	\centering
	\includegraphics[width=\linewidth]{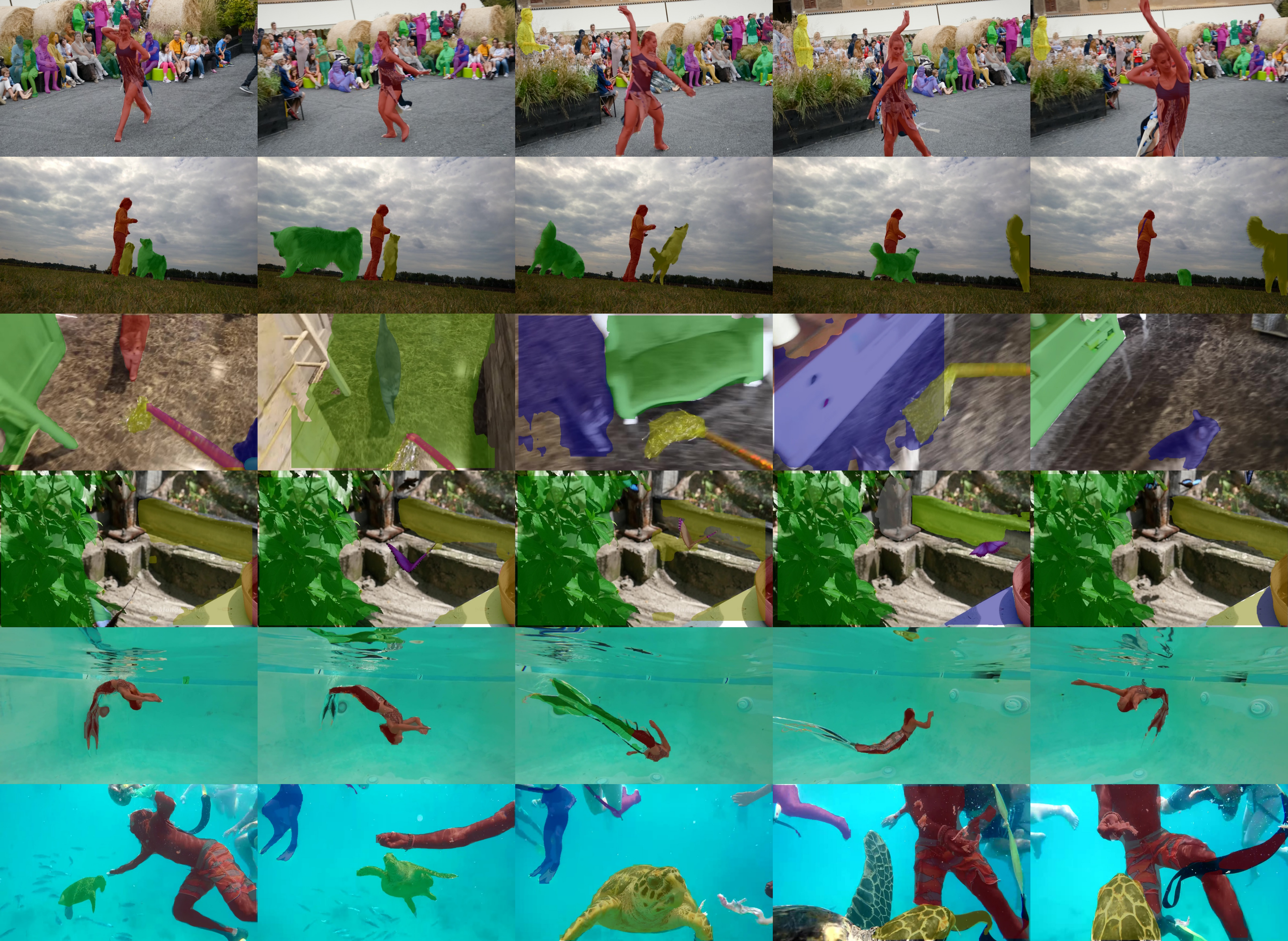}
	\caption{Additional qualitative results of UnOVOST.}
	\label{fig:qual2}
\end{figure}

Table \ref{table:results} shows our results on these three UVOS benchmarks, and compares our method to three other methods that have presented UVOS results. UnOVOST outperforms all other previous UVOS algorithms over all datasets, often by a large margin. The \texttt{val} set is significantly easier than the other two datasets, in this easier setting UnOVOST has the largest margin over the other methods. This shows that when scenes are not too crowded or complex our method does an exceptional job of successfully tracking and segmenting objects through videos. The \texttt{test-dev} set is significantly more difficult, and yet the UnOVOST algorithm can still perform extremely well, especially when compared to the performance of RVOS \cite{rvos}.  We present additional qualitative results in Figure \ref{fig:qual2}.

\PAR{Comparison to Semi-Supervised VOS Methods.}
As well as comparing to other UVOS methods, we also compare our results on the DAVIS 2017 \texttt{val} set to the current state-of-the-art semi-supervised VOS methods. Figure \ref{fig:speedplot} plots the $\mathcal{J}\&\mathcal{F}$ metric of approaches against their runtime per frame. Although these methods use the given first-frame mask, so they know exactly what objects need to be tracked in the video, our UnOVOST algorithm outperforms many of these methods, even though it operates completely unsupervised without having access to the first frame. Furthermore, many of these methods \cite{OSVOS,luiten2018premvos,voigtlaender17BMVC} extensively finetune their segmentation and tracking algorithms on the appearance of the given first-frame objects. Our method still outperforms many of these methods while being significantly faster.

\PAR{Runtime Analysis.}
In Table \ref{table:runtime} we provide detailed runtime analysis of our UnOVOST algorithm across the three datasets that we test on. The whole algorithm is able to run at around 1 frame per second (fps). The bottleneck of our algorithm is the proposal generation using Mask R-CNN which is more than 70\% of the total runtime. We use two other networks to extract features for matching these proposals over time. Our optical-flow and appearance embedding networks, while both reasonably fast at around 10 fps each, make up a combined 20\% of the runtime. The remaining runtime for actually running our algorithm, including proposal sub-selection and clipping, tracklet generation, tracklet merging with the FPC algorithm, and track saliency estimation and selection runs in around 0.07 seconds per frame, or at around 15 fps.

\begin{table}[t]
	\scriptsize
	\newcommand{\mmr}{\arrayrulecolor{lightgray}\cmidrule[0.25pt]{2-4}\arrayrulecolor{black}}
	
	\setlength{\tabcolsep}{15pt}
	\begin{center}
		\begin{tabularx}{\textwidth}{cccc}
			\cmidrule[1pt]{1-4}
			\multicolumn{1}{c}{} & {\mc{U17 \\ Val}} & {\mc{U17 \\ T-D}} & {\mc{U17 \\ T-C}} \\	
			
			\cmidrule[1pt]{1-4}
			
			Mask R-CNN           & 0.74 & 0.78 & 0.77 \\
			Optical Flow         & 0.10 & 0.14 & 0.12 \\
			Appearance Embedding & 0.10 & 0.15 & 0.11 \\
			UnOVOST Tracking     & 0.08 & 0.07 & 0.06 \\
			\cmidrule[1pt]{1-4}
			Total        & 1.02 & 1.15 & 1.06 \\
			
			\cmidrule[1pt]{1-4}
			
		\end{tabularx}
	\end{center}
	\caption{Runtime analysis of UnOVOST on the DAVIS 2017 Unsupervised \texttt{val}, \texttt{test-dev} and  \texttt{test-challenge} datasets. Times are seconds per frame.} 
	\label{table:runtime}
\end{table}

\begin{figure}[t!]
	\centering
	\includegraphics[width=\linewidth]{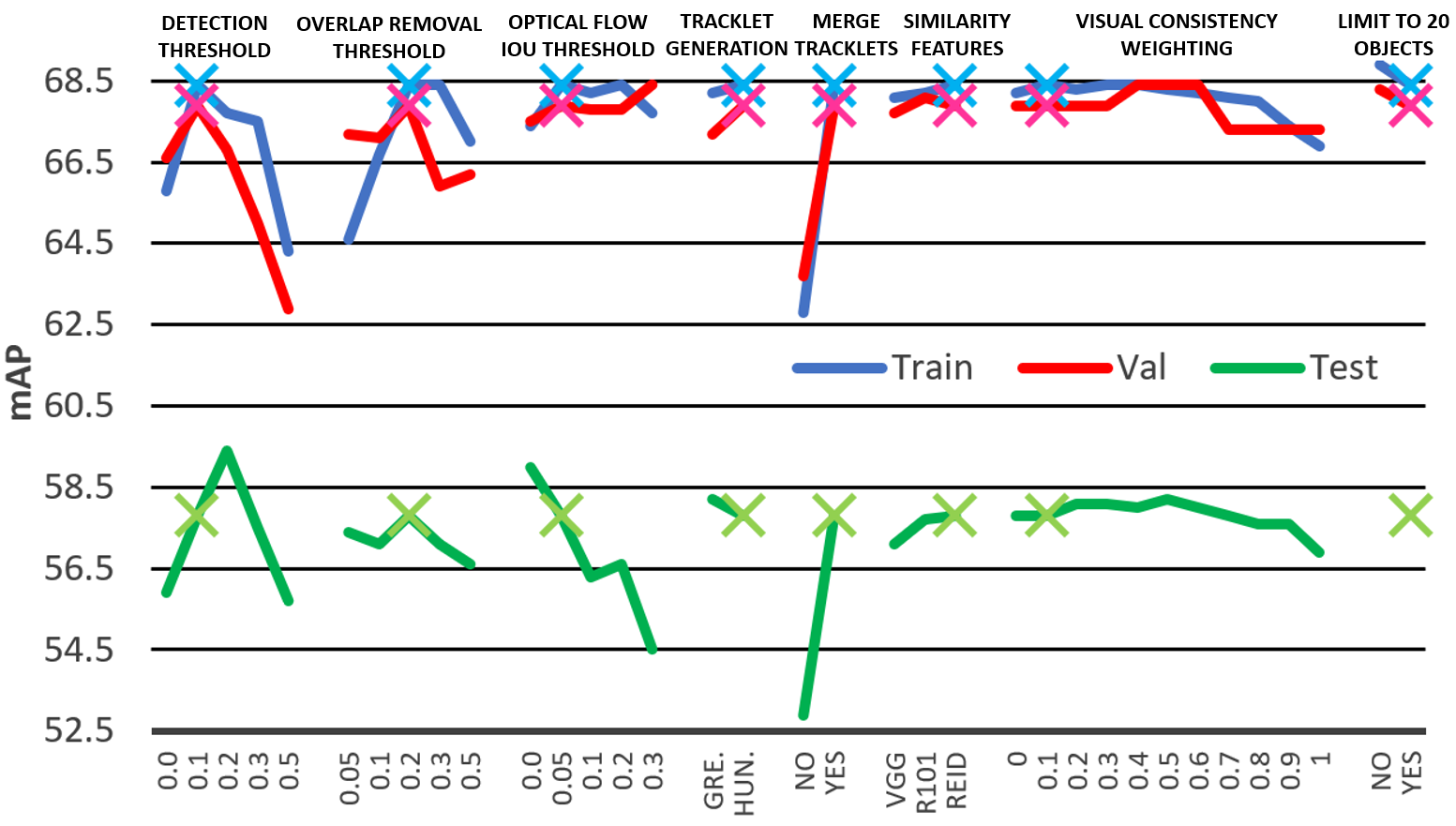}
	\caption{Results of ablating a number of design decisions for UnOVOST on the three splits of the DAVIS 2017 Unsupervised dataset. A cross indicates the selected option chosen as the best performing option on the training set.}
	\label{fig:ablation}
\end{figure}

\PAR{Ablation of the Method.}
We perform an extensive ablation of all design decisions for UnOVOST, for which the results can be found in Figure \ref{fig:ablation}. For all design decisions we use the method which performs best on the training set, even though this is often not the best on the other data splits, except for limiting the output to 20 objects which is required by the evaluation. Interestingly across all experiments, results on \texttt{train} and \texttt{val} are very similar, whereas \texttt{test-dev} is much harder, and often shows different trends in the results than the other two sets. 

First we ablate different Mask R-CNN threshold values for our input proposals. With too small or too large a threshold performance degrades significantly. We then ablate the threshold for which to remove proposals if their masks overlap. Again it is important to select a reasonable threshold. Next we ablate the IoU threshold required for merging proposals into tracklets, which is not so important for the easier validation set, but has large effects on the test set results. We use two different tracklet merging strategies, either using the Hungarian algorithm for associating proposals into tracklets or alternatively using greedy merging. Hungarian performs better on the train and validation splits but worse on the test split. We show resulting scores when evaluating just our tracklets from the first stage, compared to merging these tracklets in a second stage. Across all sets the second stage is incredibly important. We ablate the use of different similarity features for comparing tracklets, as well as the ReID vectors we compare to using last layer activations of pretrained ResNet 101 \cite{He_2016} and VGG \cite{simonyan2014very} models trained on ImageNet. The ReID vectors outperform the other similarity features on the train and test set, but the ResNet 101 features perform slightly better on the validation set. We ablate different weightings of the Visual Consistency Score and the Temporal Density Score used in our algorithm. Interestingly, the $90$:$10$ ratio which we use as it slightly outperforms other ratios on the training set, is far from the optimal weighting on the other sets. A $50$:$50$ weighting of the two scores performs the best on both the validation and testing set, with more heavily relying on either of scores performing worse. Finally we test the result of our algorithm if we relax the constraint that we can only select 20 objects for evaluation (we could not do this for the hidden test test). Our algorithm performs slightly better in this setting indicating that this restriction removes some correct objects. We recommend that UnOVOST to be used without this restriction step when being applied in the wild, and consider this only an adaption to the dataset and evaluation.

\PAR{Extension to Video Instance Segmentation.}
The task of Video Instance Segmentation (VIS) is very similar to UVOS, however in VIS the objects to be tracked must be classified into a set of predefined classes rather than just being salient throughout a video. To investigate the generalization of UnOVOST beyond the UVOS domain we run our algorithm on the YouTube-VIS dataset \cite{VIS} after training our detector and segmentor on the set of 40 classes in this dataset and adding another classification network to improve classification results. Apart from that we run UnOVOST with exactly the same parameters as for the unsupervised DAVIS task. Details of how we trained our detector and segmentor for VIS, and of the classifier we used can be found in the supplemental material.

The VIS task is evaluated using the mAP metric. This is similar to the mAP metric used for instance segmentation \cite{lin2014microsoft}, however it has been extended from single images to video. Details of mAP for VIS can be found in \cite{VIS}.

The previous state-of-the-art VIS method is MaskTrack R-CNN \cite{VIS}, which achieves a mAP scores of $30.3$ and $32.2$ on the YouTube-VIS validation and test set respectively. UnOVOST significantly outperforms this, achieving mAP scores of $44.8$ and $46.7$ on the two benchmarks respectively. With these scores UnOVOST also won the 2019 YouTube-VIS Challenge on Video Instance Segmentation, outperforming 18 other methods. In the supplementary material we present a table comparing results to all previous benchmarks and 2019 challenge entries.

\vspace{-4pt}

\section{Conclusion}
\vspace{-3pt}
In this paper, we present the novel UnOVOST (Unsupervised Offline Video Object Segmentation and Tracking) algorithm for tackling the unsupervised video object segmentation task. Our algorithm is able to track and segment a huge variety of objects in complex scenes by combining both spatio-temporal consistency and visual similarity cues in a novel tracklet based Forest Path Cutting algorithm for performing data association. UnOVOST outperforms all previous UVOS methods, while even performing competitively with many semi-supervised video object segmentation algorithms without requiring any human input as to which objects should be tracked and segmented.

{\footnotesize \PAR{Acknowledgments:} This project has been funded, in parts, by ERC Consolidator Grant DeeViSe (ERC-2017-COG-773161) and by a Google Faculty Research Award. We would like to thank Paul Voigtlaender for helpful discussions.}

\newpage
\vskip .375in
\twocolumn[\begin{center}
{\Large \bf Supplemental Material: UnOVOST: Unsupervised Offline Video Object Segmentation and Tracking \par}
\vskip 1.5em
\end{center}]
\appendix 
\section{Adapting UnOVOST to Video Instance Segmentation}

We adapt UnOVOST to the Video Instance Segmentation (VIS) domain is the following way.
\PAR{Detection.}
For detection we adapt the Mask R-CNN \cite{He17ICCV} detector to the YT-VIS benchmark to detect the 40 object classes.

To adapt this network to VIS, we created a training set by combining the YT-VIS \cite{VIS}, COCO \cite{lin2014microsoft} and OpenImages \cite{openimages} datasets.
We trained this detector on 39 classes, the 40 classes of YT-VIS with ``monkey" and ``ape" combined. This is because OpenImages only has a class which is a mix, and because in the YT-VIS training set it is unclear exactly what the difference between these two classes should be (e.g. baboons are labeled as both ape and monkey, some gorillas mislabeled as monkeys). Thus we detect these classes together and rely on our classifier later to distinguish between the two.

For COCO we use the 19 classes which overlap with the YT-VIS classes. The ``bird" class was set to ignore regions (as multiple birds such as owl, eagle and duck are in YouTube-VIS).
We map the OpenImages classes to YouTube-VIS classes, with all of our 39 classes being mapped to by at least one OpenImages class. We only use images that contain at least one annotation from our 39 classes that is not a person (because of too many people in OpenImages). We set all of the background of OpenImages images to be ignore regions and we don't sample negatives from this dataset (as OpenImages is not densely annotated).
We reweight how often we sample each image during training for class balancing. Classes are sampled such that in one epoch there are at least 5000 examples of each class. This results in sharks being sampled 18 times more often than horses. Also images form the YT-VIS dataset are sampled three times more often than those in COCO and OpenImages.

\PAR{Classification.}
The classification branch our Mask R-CNN detector works reasonably well, but still often misclassifies examples. To improve this, we use a ResNeXt-101 32x48d classifier \cite{resnext} pretrained on 940 million Instagram images \cite{wsl}, before being trained on ImageNet \cite{deng2009imagenet}. We then defined a mapping of ImageNet (INet) classes to YT-VIS classes.

This mapping results in 310 of the 1000 INet classes being mapped to our 40 YT-VIS classes, with 123 INet classes being mapped to dog and 20 to truck. Some classes are not represented (person, skateboard, giraffe, hand and surfboard). Some INet classes are mapped to multiple YT-VIS classes, e.g. ``Amphibious vehicle" being mapped to both boat and truck. There are 11 INet classes mapped to just monkey, 2 to just ape and 7 to both due to the ambiguity in YT-VIS as to what is a ape and what is a monkey.

The final INet classification score for each YT-VIS class is then the sum of the classification scores for all of the contributing INet classes.

The final classification scores were then a weighted combination of the scores from our Mask R-CNN detector and our INet trained classifier.

\PAR{Segmentation.}
We finetune the segmentation head of Mask R-CNN on the YT-VIS dataset separately for the 40 classes.

\PAR{Tracking.}
We use UnOVOST exactly as in the main paper for unsupervised VOS with exactly the same parameters. The only difference is that different input proposals are input to UnOVOST.

\begin{table}[t!]
\footnotesize
\begin{tabular}{|c|c|c|c|c|c|}

\hline 
 & mAP & AP50 & AP75 & AR1 & AR10\tabularnewline
\hline 
Ours & \textbf{46.7} & \textbf{69.7} & \textbf{50.9} & \textbf{46.2} & 53.7\tabularnewline
\hline 
foolwood & 45.7 & 67.4 & 49 & 43.5 & 50.7\tabularnewline
\hline 
bellejuillet & 45 & 63.6 & 50.2 & 44.7 & 50.3\tabularnewline
\hline 
linhj & 44.9 & 66.5 & 48.6 & 45.3 & \textbf{53.8}\tabularnewline
\hline 
minmingdii & 44.4 & 68.4 & 48.7 & 43.6 & 50.8\tabularnewline
\hline 
xiAaonice & 40 & 57.8 & 44.9 & 39.6 & 45.2\tabularnewline
\hline 
guwop & 40 & 60.8 & 43.9 & 41.2 & 49.1\tabularnewline
\hline 
exing & 39.7 & 62.1 & 42.6 & 41.4 & 46.1\tabularnewline
\hline 
MaskTrack R-CNN \cite{VIS} & 32.3 & 53.6 & 34.2 & 33.6 & 37.3\tabularnewline
\hline 
\end{tabular}
\vspace{5pt}
\caption{Results in the 2019 YouTube-VIS Challenge, compared to top 8 other participants, and the previous state-of-the-art.}
\label{table:chal}
\end{table}

\PAR{Putting it all together.}
In VIS segmentations are allowed to overlap, thus when we are not sure which class a track belongs to we propose the existence of the same track multiple times with different classes and scores. 

To obtain a track's score for each class, we average the class scores for the mask in each timestep. Frames with no masks are given 0 score thus short tracks are down weighted. We do this for both detection scores and INet scores. The final score is the weighted average of these two scores (with equal weighting). We output each track multiple times for every class with a score greater than 0.0001.
Note that the detector doesn't discriminate between apes and monkey, so the one detection score is used for both. Also our INet classifier doesn't give scores for 5 of the 40 classes, so for these we only use detector scores.

\newpage
{\small
\bibliographystyle{ieee}
\bibliography{abbrev_short,egbib}
}

\end{document}